\documentclass[letterpaper, 10 pt, conference]{ieeeconf} % Comment this line out if you need a4paper
\usepackage{subcaption}
\usepackage{graphicx}
\usepackage{changepage}
\usepackage{booktabs}
\usepackage{afterpage}
\usepackage{balance}
\usepackage{float}
\usepackage{amssymb} 
\usepackage{multirow}
\usepackage{diagbox}

\IEEEoverridecommandlockouts % This command is only needed if 
\usepackage{amsmath} % assumes amsmath package installed

\title{\LARGE \bf
MAGNNET: Multi-Agent Graph Neural Network-based Efficient Task Allocation for Autonomous Vehicles with Deep Reinforcement Learning
}

\author{Lavanya Ratnabala, Aleksey Fedoseev, Robinroy Peter, and Dzmitry Tsetserukou% <-this % stops a space
\thanks{Research reported in this publication was financially supported by the RSF-DST grant No. 24-41-02039. The authors are with the Intelligent Space Robotics Laboratory, CDE, Skolkovo Institute of Science and Technology, Bolshoy Boulevard 30, bld. 1, 121205, Moscow, Russia. 
{\tt \{lavanya.ratnabala, aleksey.fedoseev, robinroy.peter, d.tsetserukou\}@skoltech.ru}}
}% <-this % stops a space
\begin{document}

\maketitle
\thispagestyle{empty}
\pagestyle{empty}

%%%%%%%%%%%%%%%%%%%%%%%%%%%%%%%%%%%%%%%%%%%%%%%%%%%%%%%%%%%%%%%%%%%%%%%%%%%%%%%%

\begin{abstract}
This paper addresses the challenge of decentralized task allocation within heterogeneous multi-agent systems operating under communication constraints. We introduce a novel framework that integrates graph neural networks (GNNs) with a centralized training and decentralized execution (CTDE) paradigm, further enhanced by a tailored Proximal Policy Optimization (PPO) algorithm for multi-agent deep reinforcement learning (MARL). Our approach enables unmanned aerial vehicles (UAVs) and unmanned ground vehicles (UGVs) to dynamically allocate tasks efficiently without necessitating central coordination in a 3D grid environment. The framework minimizes total travel time while simultaneously avoiding conflicts in task assignments. For the cost calculation and routing, we employ reservation-based A* and R* path planners. 

Experimental results revealed that our method achieves a high 92.5\% conflict-free success rate, with only a 7.49\% performance gap compared to the centralized Hungarian method, while outperforming the heuristic decentralized baseline based on greedy approach. Additionally, the framework exhibits scalability with up to 20 agents with allocation processing of 2.8 s and robustness in responding to dynamically generated tasks, underscoring its potential for real-world applications in complex multi-agent scenarios.
\\

\end{abstract}

%%%%%%%%%%%%%%%%%%%%%%%%%%%%%%%%%%%%%%%%%%%%%%%%%%%%%%%%%%%%%%%%%%%%%%%%%%%%%%%%
\section{Introduction}

The deployment of multi-agent systems (MAS) comprising autonomous robots and drones has witnessed remarkable growth across diverse applications, including disaster response, agricultural monitoring, surveillance, and warehouse logistics. The inherent advantages of enhanced scalability, distributed operations, and parallel execution render MAS highly appealing for addressing complex real-world challenges. However, significant bottlenecks persist in effectively assigning tasks to the most suitable agents, particularly in scenarios where tasks emerge dynamically and agents experience constrained communication capabilities.
\begin{figure}[h]
 \centering
 \includegraphics[width=0.95\linewidth]{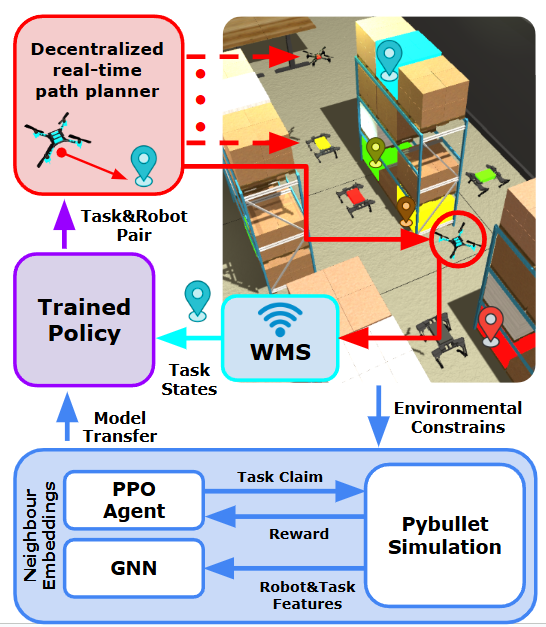}
 \caption{Multi-vehicle task allocation scenario with heterogeneous swarm.} %with communication range of one UAV represented by circle.
 \vspace{-0.5cm}
 \label{fig:cover}
\end{figure}

Traditional task allocation methods can be broadly categorized into centralized and decentralized approaches. Centralized methods, for example, the Hungarian algorithm \cite{Kuhn_1955}, offer the advantage of producing globally optimal task assignments; however, they require continuous communication of each agent's state to a designated central node. This dependency introduces several critical limitations, e.g., communication overhead, restricted scalability, and susceptibility to single points of failure. In contrast, decentralized methods exhibit greater robustness against agent failures and communication disruptions while also offering improved scalability as the number of agents increases \cite{sin_vaish}, but often yield suboptimal solutions. This can result in conflicts or idle tasks, especially under conditions of high partial observability.

To address these challenges, we introduce MAGNNET, an innovative framework that synergizes multi-agent deep reinforcement learning (MARL) with graph neural networks (GNNs) within a CTDE paradigm. During the training phase, a global critic observes the comprehensive state of the system, subsequently updating the agents' policies using PPO algorithm to facilitate global optimization. In the execution phase, agents operate independently, relying exclusively on local observations, augmented by GNN-enhanced message passing. The GNN plays a pivotal role in shaping relational embeddings for each agent by analyzing local neighborhood information and agent-to-task interactions, which empowers agents to make informed decisions regarding the acceptance or rejection of incoming tasks.

% Experimental results obtained from simulations in a 3D PyBullet-based environment indicate that \emph{ MAGNNET} achieves near-optimal performance in comparison to centralized methods while significantly outperforming greedy and random baselines in minimizing overall costs and avoiding conflicting assignments. Moreover, the system maintains high success rates free from conflicts, even as the number of agents and tasks increases, demonstrating the scalability of our approach in larger systems.

The proposed framework can extend beyond UAVs and UGVs, adapting to facilitate autonomous vehicle taxi matching, thereby reducing passenger waiting times while preventing simultaneous pickups by multiple vehicles.

\section{Related Works}

Multi-agent task allocation (MATA) has been extensively studied in the fields of robotics, autonomous systems, and artificial intelligence. Existing approaches can be broadly categorized into optimization-based methods, market-based mechanisms, and learning-based approaches. Furthermore, distinctions between centralized and decentralized task allocation, as well as homogeneous and heterogeneous multi-agent settings, are critical in determining the efficiency and adaptability of such systems.

\subsection{Optimization-Based Approaches}
Optimization techniques, e.g., the linear sum assignment problem (LSAP) and the Hungarian algorithm, have been widely applied in multi-agent task allocation \cite{goarin2024graph}. While these methods provide optimal solutions, they often rely on centralized computation, rendering them unsuitable for large-scale real-time applications. Ismail et al. \cite{ismail2017decentralized} proposed a novel decentralized-based Hungarian method to solve this problem. Another decentralized version of the Hungarian method is proposed by Xia et al. \cite{xia} to solve task allocation for underwater vehicles. These methods extend the Hungarian approach to decentralized settings by ensuring task allocation remains optimal as long as agent networks remain connected. Kong et al. \cite{greedy} proposed a new optimization strategy that combines the improved particle swarm optimization and the greedy IPSO-G algorithm. Moreover, two different stochastic approaches, the Genetic Algorithm (GA) and the Ant-Colony Optimization (ACO) algorithm, were introduced by \cite{genetic} to solve the multi-robot task allocation problem. Peter et al. \cite{neurofleets} introduced a swarm intelligence-based task allocation method utilizing decentralized decision-making and subgoal-based path formation.

\subsection{Market-Based Mechanisms for Task Allocation}
Market-based strategies, particularly auction-based methods, have been widely explored for decentralized task allocation. Zhong et al. \cite{zhong2018stable} proposed an extended auction-based driver-passenger matching system for ride-sharing applications. The system optimizes both stability and fairness in assignments by allowing drivers to bid on passengers based on individual profitability. Liu et al. \cite{liu2024multi} extended auction-based methods to multi-UAV systems, incorporating Dubins path-based flight cost estimation to refine task allocations. While auction-based strategies provide high adaptability, their efficiency heavily depends on accurate bid valuation models. An alternative hybrid approach is introduced in the Harmony Drone Task Allocation (DTA) method \cite{harmonyDTA}, which integrates a consensus-based auction mechanism with a gossip-based consensus strategy. The Harmony DTA optimizes multi-drone task allocation under complex time constraints by balancing task urgency with resource availability while minimizing communication load.

\subsection{Multi-Agent Deep Reinforcement Learning (DRL) in Task Allocation}

The integration of DRL into MATA has led to promising advancements, particularly through MARL frameworks. Agrawal et al. \cite{agrawal2023rtaw} introduced an attention-inspired DRL method for warehouse-based task allocation that optimizes decision-making efficiency and scalability. Moreover, recent studies explore Multi-Agent Proximal Policy Optimization (MAPPO) to enhance collaborative decision-making and mitigate non-stationary challenges in multi-agent environments \cite{mappo2023arxiv}. Additionally, it incorporates message pooling and weight scheduling mechanisms to enhance agent communication, improving efficiency in cooperative decision-making. Lowe et al. \cite{lowe2017multi} use Q-learning to do cooperative and competitive tasks. QMIX is a CTDE approach that shows \cite{rashid2018qmix} promising results in cooperative tasks.

\subsection{Graph Neural Networks (GNNs) for Decentralized Task Allocation}
Recent advances in graph-based learning methods have demonstrated substantial potential for improving decentralized task allocation. GNNs enable robots to model and process inter-agent relationships, allowing task allocations to be computed based on local graph structures rather than centralized control. Goarin et al. \cite{goarin2024graph} proposed DGNN-GA, a decentralized GNN-based goal assignment method that optimizes communication efficiency in multi-robot planning. Heterogeneous multi-agent systems pose additional complexities due to differences in robot capabilities, sensor modalities, and mobility constraints. The GATAR framework \cite{peng2024graph} introduced a graph-based task allocation framework for multi-robot target localization, specifically designed for heterogeneous robot systems. Furthermore, Blumenkamp et al. \cite{blumenkamp2022framework} develop a real-world framework for deploying decentralized GNN-based policies in multi-robot systems, facilitating seamless sim-to-real transfers. 

Despite advancements in MATA, several challenges remain unaddressed or partially solved. Many DRL-based solutions struggle to generalize to larger teams due to the inherent non-stationarity of multi-agent environments. While GNN-based models enhance decentralized coordination, ensuring robust communication under limited bandwidth conditions remains a challenge. The transition from simulation to real-world applications is hindered by the lack of standardized multi-agent benchmarks and sim-to-real transfer limitations. 
 Future research should focus on hybrid approaches that combine optimization, DRL, and GNNs to balance efficiency, adaptability, and real-time feasibility in multi-agent decentralized task allocation.

\section{Multi-Vehicle Task Allocation Approach}

\subsection{Problem Formulation}
We consider $N$ agents (or robots), denoted by $\{a_1,a_2,\dots,a_N\}$, and $M$ tasks, $\{T_1,T_2,\dots,T_M\}$. Each task $T_j$ is characterized by a location in $\mathbb{R}^3$, denoted $\mathbf{l}_j$, and a status such as \emph{Waiting} or \emph{Assigned}. Each agent $a_i$ has a position $\mathbf{p}_i\in \mathbb{R}^3$ and a status. Status are categorized as \emph{idle} which means that agent does not act; \emph{accept} which means it is willing to accept the task; \emph{assign} is the final decision to assign the task; once they complete the task, state change to \emph{complete}.

The $c_{ij}$ is the cost for $a_i\text{ to complete }T_j$, i.e., travel time to complete the task. Then $x_{ij}\in\{0,1\}$ is the combinatorial variable, where $x_{ij}=1$ implies $a_i$ is assigned to $T_j$. In our approach, we don't directly solve $x_{ij}$; rather, each agent selects an action (request or reject) for any given task. Thus, the discrete action space for each agent $a_i$ is denoted as:
\begin{equation}
 A_i \;=\; \{\,0,1,\ldots,M\},
\end{equation}
where “0” is the action of rejecting the task and $j$ is the integer within $1 \le j \le M$ that indicates “requesting task $T_j$". If multiple agents request the same $T_j$ within one decision step, the environment assigns $T_j$ to whichever agent has the smallest cost $c_{ij}$. The overall objective is to minimize $\sum_{i=1}^N\sum_{j=1}^M c_{ij}\,x_{ij}$, subject to each task being assigned to at most one agent. Because this objective is inherently global, a centralized mechanism is beneficial during training to evaluate team-level performance. The environment is modeled as a heterogeneous graph $\mathcal{G}$ consisting of two node types.

\noindent
\emph{Agent Nodes} $\{a_1,\dots,a_N\}$: Each agent node stores position $\mathbf{p}_i \in \mathbb{R}^3$, agent status, and cost estimates $\{\,c_{ij}\}$ for all available tasks $T_j$. Cost is basically total travel time. We will explain how we calculate this cost in upcoming sections.

\noindent
\emph{Task Nodes} $\{T_1,\dots,T_M\}$: Each task node stores position $\mathbf{l}_j \in \mathbb{R}^3$ and task status.

\noindent
\emph{Agent--Task Edges}: Each edge $(a_i, T_j)$ exists between agent $a_i$ and task $T_j$ to allow the agent to evaluate its suitability of completing the task. In our case, every agent node is connected to every task node, ensuring that each agent has access to global task information. This fully connected structure allows agents to make informed choices based on cost, availability, and competition, ultimately improving the efficiency of decentralized task allocation. Multi-layer convolutional graph neural network (GNN) is applied to transform node and edge features into final \emph{embeddings}. After $L$ message-passing iterations, the embedding for agent node $a_i$ is given by:
\begin{equation}
 \mathbf{z}_i^{(L)} \;=\;
 \mathrm{GNN}\Bigl(\mathbf{z}_i^{(0)},\,\{\mathbf{z}_k^{(0)} : k \in \mathcal{N}(i)\},\,E\Bigr),
\end{equation}
where $\mathbf{z}_i^{(0)}$ is the initial feature vector of the node $a_i$, $\mathcal{N}(i)$ is the set of its neighbors, and $E$ is the edge attributes. Each agent then incorporates $\mathbf{z}_i^{(L)}$ alongside its other local observations when selecting actions.

\subsection{Centralized Training \& Decentralized Execution in Multi-Agent Deep Reinforcement learning}

\subsubsection{Centralized Training}
Minimizing the total cost $\sum c_{ij}$ is a global objective, requiring coordination among agents. However, since each agent only has access to local observations, it cannot independently evaluate how well its decision contributes to overall team performance. To address this, we employ a centralized critic (or value network) $V_{\text{central}}^\pi(\cdot)$ during training, which has access to the full environment state, including all agents' positions, statuses, and task assignments. The critic provides a global value estimate that helps guide each agent’s individual policy updates toward minimizing the overall cost. 

Importantly, while the critic is centralized, the decision-making process remains fully decentralized throughout training. Each agent still selects actions based only on its own local observations and GNN-processed independent information. The centralized critic is used solely for computing better advantage estimates and updating policies more effectively. Once training is complete, the critic is discarded, and agents execute their policies in a fully decentralized manner, relying only on their individual learned strategies.

\subsubsection{Decentralized Execution}
At execution time, each agent $a_i$ maintains its own learned policy $\pi_i$, which takes observations of local features (agent status, cost details to available tasks) and the GNN embedding $\mathbf{z}_i^{(L)}$ as input, summarizing the local-neighbor and task details.
This policy outputs a discrete action in $\{0,\dots,M\}$ indicating whether to \emph{reject} (0) or \emph{accept} (some $T_j$). Because each agent's policy runs independently based on localized data, no centralized coordinator is needed at inference time. This preserves scalability and autonomy in dynamic environments.

\subsection{Model Architecture and Training Strategy}

\subsubsection{Neural Network Architecture} 
\begin{figure*}[tp]
 \centering
 \smallskip
 \includegraphics[width=0.8\textwidth]{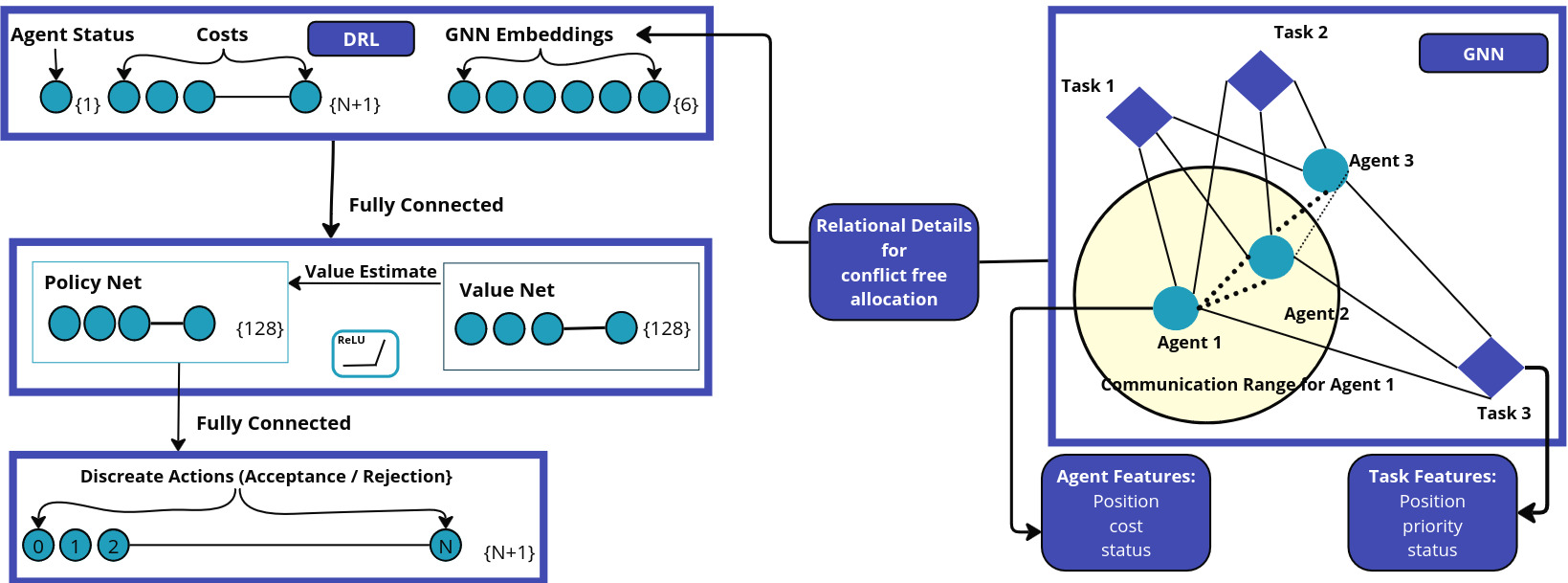}
 \caption{Multi-agent reinforcement learning architecture for task allocation.}
 \label{fig:land}
\end{figure*}
The GNN consists of two Graph Convolutional layers, each with 6 hidden units. These layers enable agents to aggregate information from neighboring agents within a communication range and from task nodes, facilitating more informed task allocation decisions. Each layer is followed by a ReLU activation function to introduce non-linearity and improve representational capacity. The input features to the GNN include agent positions, agent statuses, cost estimates for available tasks, and task positions. After message passing, the final agent embeddings $\mathbf{z}_i^{(L)} \in \mathbb{R}^6$ are obtained, encoding both local information and relational dependencies.

Each agent's policy network is a multi-layer perceptron (MLP) that takes as input the concatenation of $\mathbf{z}_i^{(L)}$ with local observations, producing a probability distribution over the discrete action space $\{0,\dots,M\}$. The input layer has a dimension of $M+7$, where $M+1$ corresponds to local observations consisting of the agent's status and cost estimates for $M$ tasks, while the remaining six dimensions correspond to the GNN-generated embedding. The network contains a hidden layer with 128 neurons, followed by a ReLU activation, and an output layer that applies a softmax activation to produce a probability distribution over available actions.

The central value network follows a similar MLP architecture but processes a global representation of the environment. It takes as input the concatenation of all agent embeddings and task states, capturing the full system dynamics. The network includes a hidden layer with 128 neurons followed by a ReLU activation and outputs a single scalar value estimate, representing the overall value of the system state. The policy and value networks share feature representations and are optimized using the PPO algorithm. The overall neural network architecture is shown in Fig.~\ref{fig:land}.

\subsubsection{Learning Algorithm} 
We developed an algorithm based on the PPO approach of Ray RLlib, a scalable reinforcement learning framework designed for distributed multi-agent training. The policy and value networks are optimized using the Adam optimizer with a learning rate of $1 \times 10^{-5}$. Training is conducted with a batch size of 512, using a stochastic gradient descent (SGD) mini-batch size of 64 and 10 SGD iterations per update. To encourage exploration and prevent premature convergence, an entropy coefficient of $0.05$ is applied. The discount factor is set to $\gamma = 0.99$ to ensure that agents optimize long-term task allocation efficiency. Generalized advantage estimation (GAE) with $\lambda = 0.95$ is employed to reduce variance while maintaining bias in advantage computation, leading to more stable training updates.

% Training follows a complete-episode rollout strategy within RLlib’s multi-agent environment framework. Each agent observes only its local state and relational embeddings from the GNN. However, a centralized critic evaluates the overall system state, providing advantage estimates to improve policy optimization during training. The critic allows for more effective policy updates by ensuring that individual agents align their decisions with the global objective of minimizing the total task allocation cost. Once training is complete, the centralized critic is discarded, and agents execute decisions in a fully decentralized manner based solely on their learned policies and local observations.

\subsubsection{Reward Shaping} 
 Agents receive a positive reward when they successfully request and are assigned a task for which they have the lowest cost among all the contenders. A penalty is applied when multiple agents attempt to claim the same task, ensuring that task allocation remains conflict-free. Additionally, agents receive a negative reward if they repeatedly reject tasks without a valid reason, preventing stagnation and encouraging active task participation. To ensure global efficiency in task assignments, a system-wide bonus is provided when all tasks are successfully allocated with minimal total cost. 
 % This reward structure aligns local agent decisions with the overall system objective, incentivizing cooperative behavior that collectively minimizes total travel time and optimizes team-wide task execution.

\subsubsection{Training Duration and Deployment} 
The number of training steps required for policy convergence depends on the complexity of the task allocation scenario, with stable learning typically achieved between 200k and 500k environment interactions. During training, the centralized critic leverages global state information to refine individual policies, ensuring that agents learn strategies that improve long-term task efficiency. Once training is complete, the centralized critic is discarded, and each agent operates in a fully decentralized manner. Agents rely solely on their local sensing capabilities and GNN-derived relational embeddings to make real-time autonomous decisions. The final deployment strategy ensures that task allocation remains fully decentralized, with agents selecting their actions independently.

\subsection{Simulation Setup and Path Planning}

\subsubsection{Simulation Environment}
\begin{figure}[h]
 \centering
 \includegraphics[width=1.0\linewidth]{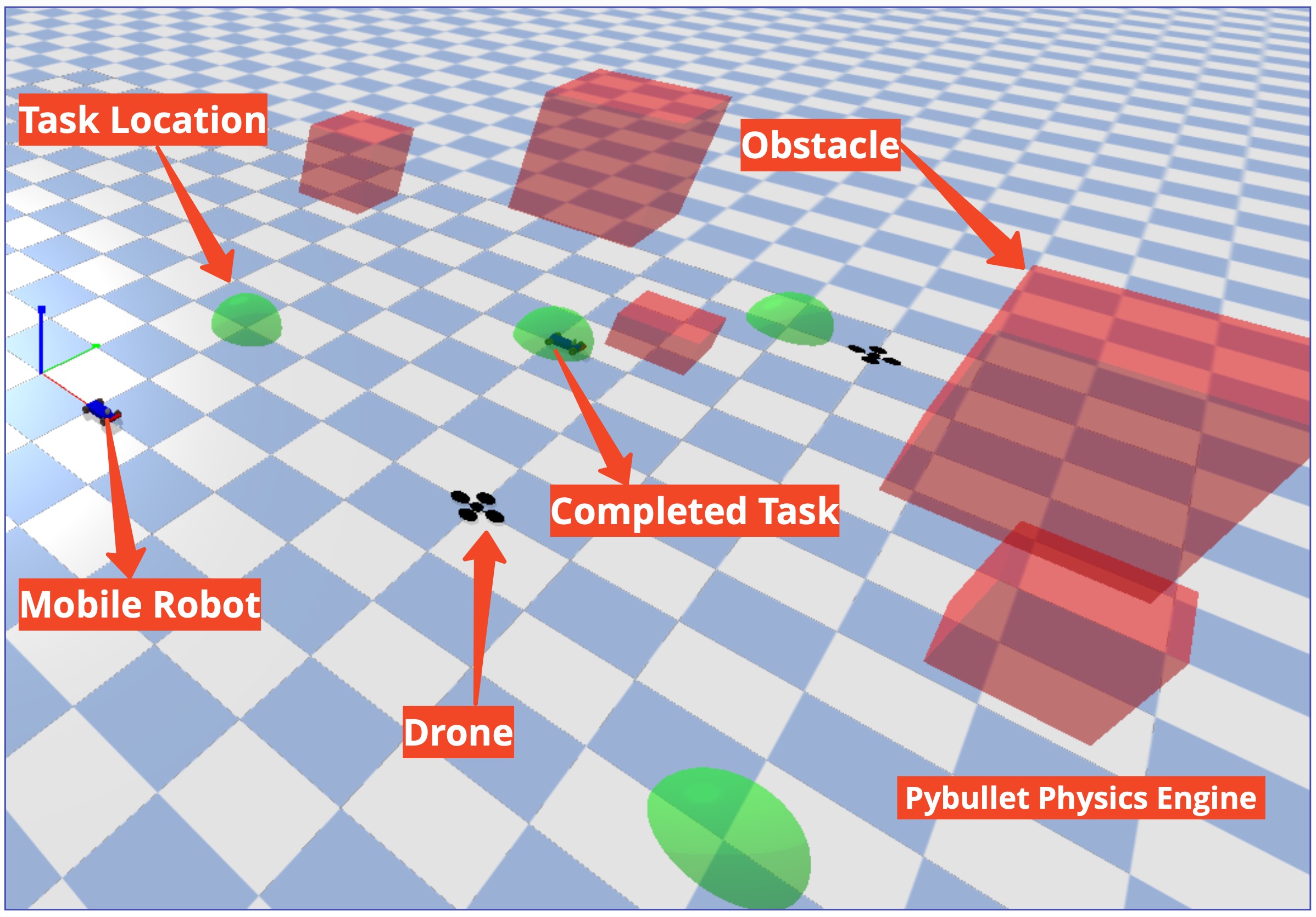}
 \caption{Simulation environment in PyBullet. Agents navigate with dynamic task assignments and obstacles.}
 \label{fig:sim_setup}
\end{figure}
We developed a custom PyBullet-based simulator to train and evaluate decentralized task allocation in a heterogeneous multi-agent system. Fig.~\ref{fig:sim_setup} shows our simulation environment. The environment is a three-dimensional grid-based world, where agents navigate to complete dynamically assigned tasks while avoiding obstacles. The simulation is integrated with Ray RLlib for reinforcement learning. 

% An overview of the parameter configuration used in the PyBullet-based custom simulator is shown in Table~\ref{tab:sim_env}.

% \begin{table}[h]
% \centering
% \caption{The Environments and Conditions of Simulation.}
% \label{tab:sim_env}
% \begin{tabular}{|l|l|}
% \hline
% \textbf{Aspect} & \textbf{Details} \\ 
% \hline
% \textbf{Grid Size} & $50 \times 50 \times 30$ (1 m per unit) \\
% \textbf{Number \& Type of Robots} & 20 UAVs and UGVs \\
% \textbf{Task Generation} & Dynamic (20 active tasks at a time) \\
% \textbf{Obstacle Density} & 10 obstacles (randomly placed) \\
% \textbf{Simulation Step Time} & 0.0005 s per step \\
% \hline
% \end{tabular}
% \end{table}

\subsubsection{Path Planning and Cost Computation}
Task allocation alone does not guarantee efficient execution; agents must navigate to their assigned tasks efficiently. We utilize an A* path planning algorithm to compute the shortest path between an agent and its assigned task, considering obstacles and other agents in the environment. The computed path is then used to calculate the task completion cost based on the agent velocity. The cost function is formulated as:

\begin{equation}
 c_{ij} = \frac{d_{ij}}{v_i}
\end{equation}
where $c_{ij}$ is the cost for the agent $a_i$ to complete the task $T_j$, $d_{ij}$ is the shortest path distance from agent $a_i$ to task $T_j$, computed with A* and $v_i$ is the velocity of the agent. Since multiple agents navigate toward their respective tasks, a reservation-based conflict resolution mechanism is applied to prevent collisions. Prior to moving, each agent verifies if the target position is already occupied or reserved and if an obstacle obstructs the path. In case of a conflict, an agent initiates dynamic replanning using an updated A* search. If a conflict arises, the system resolves it through priority-based resolution: the agent with the lowest task completion cost retains its path, while others replan. 
%Alternative Path Search: The conflicting agent recomputes a new path using the A* algorithm. Wait-and-Proceed Strategy: 
If no alternative path exists, the agent waits for the occupied space to become available. 
This combination of A* for distance computation, velocity-based cost estimation, and conflict-aware navigation ensures efficient and conflict-free decentralized task execution.

\section{Experiments}
\subsection{Experimental Setup}

\label{sec:experiments}
% \begin{figure*}[t]
% \centering
% \includegraphics[width=0.65\textwidth]{figures/setup.jpg}
% \caption{Experimental setup illustrating the environment, training process, agents, and task details}
% \label{fig:systemview}
% \end{figure*}
We use $50\times 50\times 30$ m environment in Pybullet, populated by $N$ agents: a mix of ground robots (velocity $\sim 3$ m/s) and drones (velocity $\sim 5$ m/s). A maximum of $20$ tasks can be active, appearing at random locations at fixed intervals. Agents plan paths around randomly placed obstacles, computing cost-to-task via $A^*$ path planner. All the experiments were conducted in simulation.

We evaluated three metrics for the algorithms: the total travel time (cost), estimated as the sum of all task travel times; the conflict-free success rate, estimated as a fraction of tasks successfully assigned to exactly one agent; and the allocation time, estimated as the time needed for the agents to reach an assignment decision.
Our method was compared against Hungarian, which gives the centralized optimal solution. Our aim was to achieve the closest possible value of the total travel cost with a centralized approach. We additionally compared MAGNNET with a heuristic Greedy-based allocation \cite{braquet2021675} and Random assignment to check how we reduced the total travel time from our allocation. The scalability of the system was tested in experiments with a number of robots from 4 to 20. 
% The overall MAGNNET model for training and testing is described in Fig.~\ref{fig:systemview}.

\subsection{Results and Discussion}
\subsubsection{Training performance}
\begin{figure}[h]
 \centering
 \includegraphics[width=0.45\textwidth]{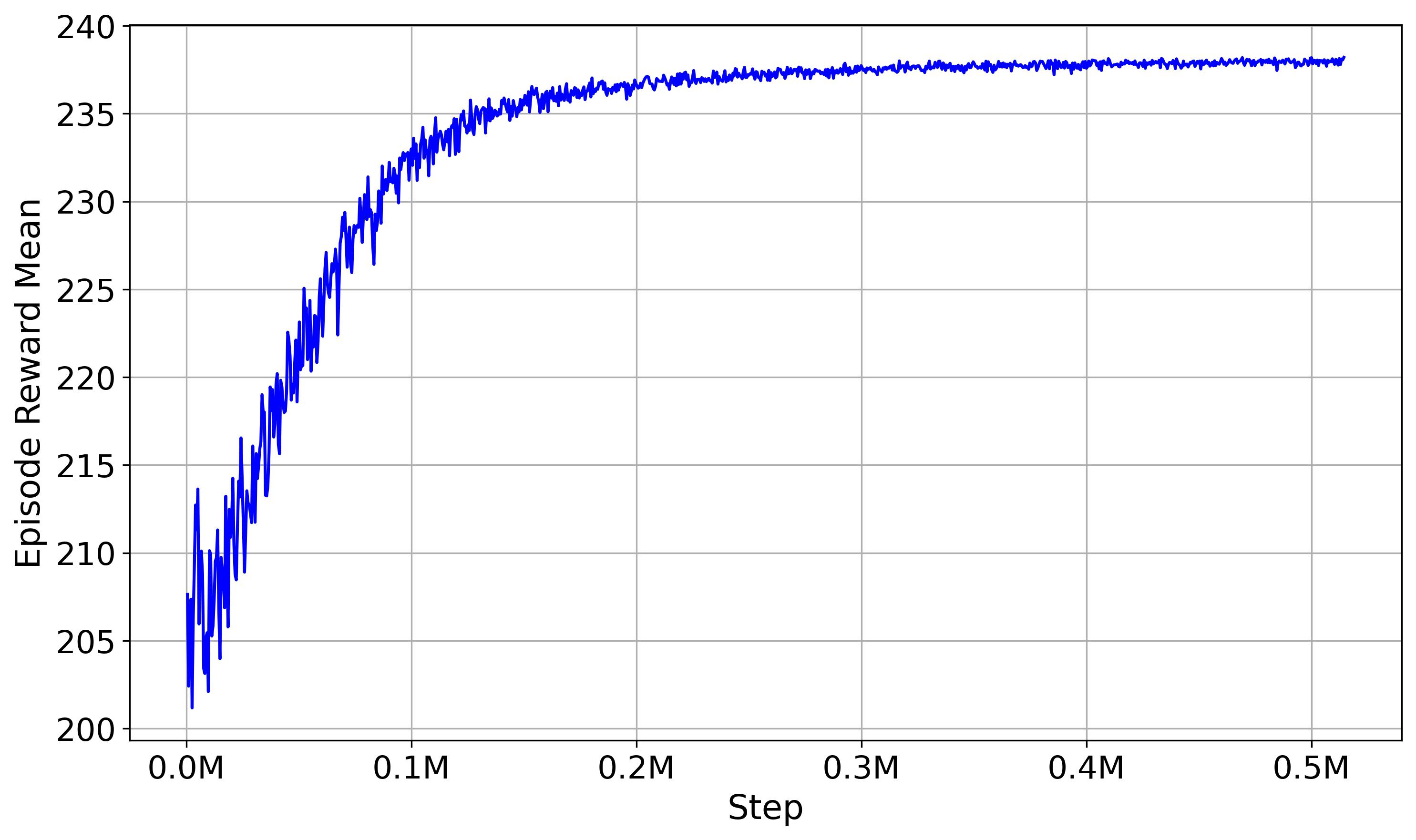}
 \caption{Mean reward vs. training steps during learning.}
 \label{fig:reward}
\end{figure}

\begin{figure}[tp]
 \centering
 \smallskip
 \includegraphics[width=0.45\textwidth]{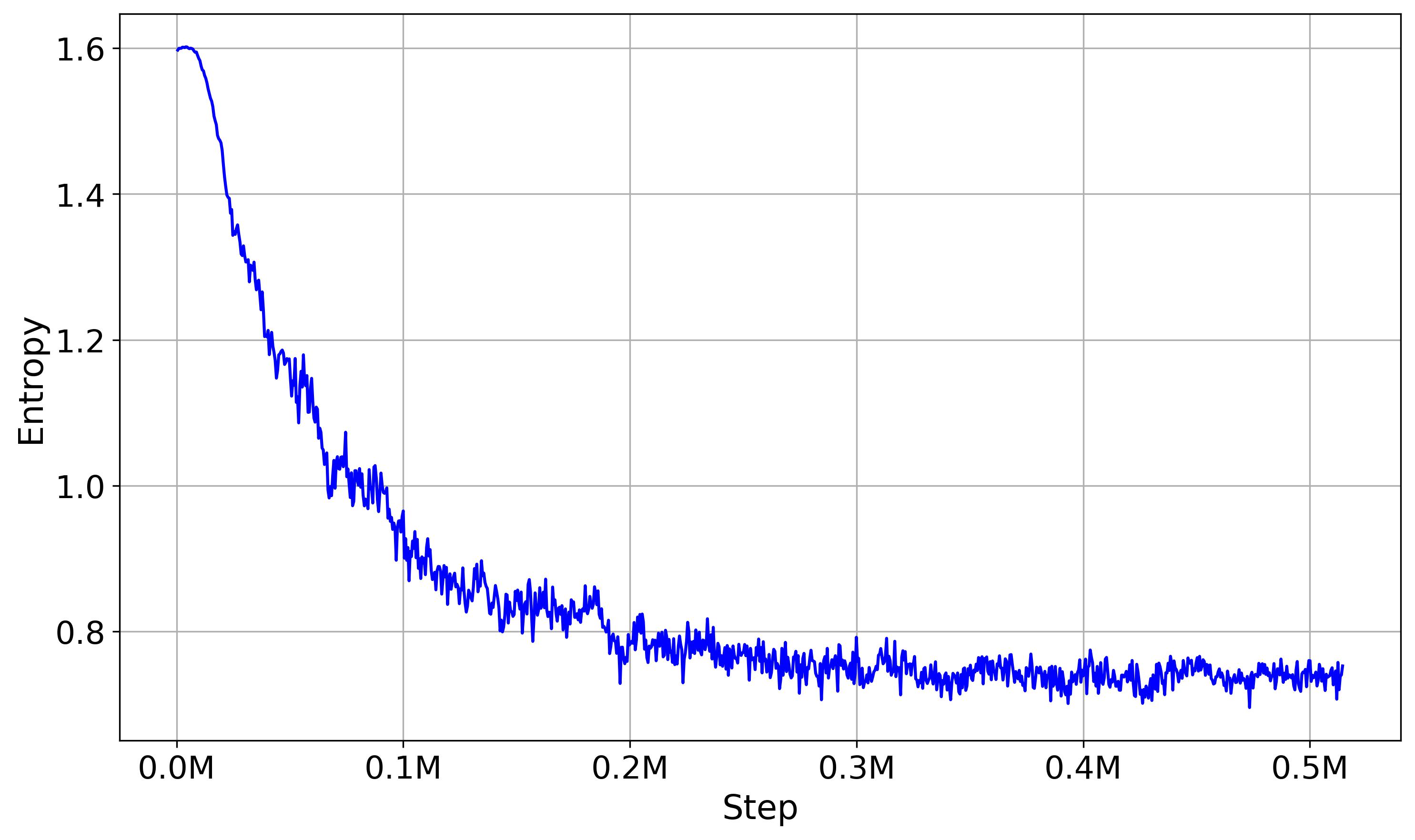}
 \caption{Entropy vs. training steps during learning.}
 \label{fig:entropy}
\end{figure}
The increase in rewards shown in Fig.~\ref{fig:reward}, coupled with a decrease in entropy shown in Fig.~\ref{fig:entropy}, indicates that agents are progressively learning to allocate tasks more efficiently over time. Initially, high entropy reflects uncertainty in decision-making due to insufficient experience, leading to suboptimal task assignments and lower rewards. However, as training progresses, agents refine their policies, resulting in more consistent and confident task selection. This is reflected in the rising reward trend, which signifies that agents are successfully maximizing their task allocation efficiency. The entropy reduction further confirms that agents are transitioning from an exploratory phase to a more deterministic and optimized decision-making process, ultimately leading to improved overall system performance.

\subsubsection{Cost of total travel time}
In the first scenario, we fix $M=N$ tasks, which appear at the start. Table~\ref{tab:comp_static} summarizes the total travel cost achieved by each method for various numbers of agents.

\begin{table}[!h]
\centering
\caption{Total Travel Cost in Static Task Scenario.}
\label{tab:comp_static}
\begin{tabular}{|c|c|c|c|c|}
\hline
{\textbf{N}} & {\textbf{Hungarian}} & {\textbf{ MAGNNET}} & {\textbf{Greedy}} & {\textbf{Random}} \\
\hline
4 & \textbf{20.3} & \textbf{20.3} & 22.5 & 27.9 \\
\hline 
8 & \textbf{60.3} & \textbf{61.2} & 65.8 & 72.3 \\
\hline 
12 & \textbf{131.9} & \textbf{134.7} & 140.5 & 175.7 \\
\hline 
20 & \textbf{254.3} & \textbf{321.5} & 383.2 & 423.8 \\
\hline
\end{tabular}
\end{table}

Since the total travel cost is invariant to the chosen path planner, we compared the performance of MAGNNET trained on reservation-based $A^*$ with a collision-free $RRT^*$ approach adapted to our discrete environment. The results of the comparison are shown in Table~\ref{tab:comp_planners}.

\begin{table}[!h]
\centering
\caption{Mean Path Length (in m) of Two Planning Algorithms Embedded in MAGNNET.}
\label{tab:comp_planners}
\begin{tabular}{|l|*{4}{c|}}\hline
\backslashbox{\textbf{Path Planning}}{\textbf{N Agents}}
&\makebox[2em]{\textbf{4}}&\makebox[2em]{\textbf{8}}&\makebox[2em]{\textbf{12}}
&\makebox[2em]{\textbf{20}}\\\hline
 $A^*$ algorithm & 5.75 & 13.63 & 20.83 & 38.40 \\\hline
 $RRT^*$ algorithm & 8.86 & 13.87 & 19.86 &40.95 \\\hline
\end{tabular}
\end{table}

The results revealed that at a lower number of agents $A^*$ outperformed the baseline method by 35\%, however, with an increasing number of agents, the performance of both algorithms is approximately equally efficient with variance within 6.5\%. 
Table~\ref{tab:comp_static} shows the comparison of the total cost versus the baseline Hungarian and decentralized the Greedy approach and random allocation. Our method approaches the centralized Hungarian baseline with an average gap of under $7.49\%$, whereas Greedy and Random show noticeably higher total cost, particularly as the number of agents grows. 
% Our method shows promising results that our decentralized MAGNNET can give optimal assignment of tasks.

\subsubsection{Conflict-free success rate}
Next, we measured the percentage of tasks allocated to a singular agent. Table~\ref{tab:success_rate} shows the conflict-free success rates and approximate allocation times. The Hungarian method achieved a 100\% conflict-free allocation expected for a centralized approach. Our decentralized approach maintains a high success rate of $80\%$ to $100\%$, significantly outperforming the Greedy algorithm, especially for larger swarm scales.

% \begin{table}[!h]
% \centering
% \caption{Conflict-free Success Rate and Allocation Time.}
% \label{tab:success_rate}
% \begin{tabular}{|c|c c c|c c c|}
% \hline
% & \multicolumn{3}{c|}{\textbf{Success Rate (\%)}} & \multicolumn{3}{c|}{\textbf{Allocation Time (s)}} \\

% \textbf{N} & \textbf{Hungarian} & \textbf{ MAGNNET} & \textbf{Greedy} & \textbf{Hungarian} & \textbf{MAGNNET} & \textbf{Greedy} \\
% \hline
% 4 & 100 & 100 & 90 & 0.8 & 0.4 & 0.3 \\
% \hline
% 8 & 100 & 100 & 80 & 1.5 & 0.6 & 0.3 \\
% \hline
% 12 & 100 & 90 & 80 & 2.8 & 1.2 & 0.5 \\
% \hline
% 20 & 100 & 80 & 60 & 5.6 & 2.8 & 1.2 \\
% \hline
% \end{tabular}

% \end{table}

\begin{table}[!h]
\centering
\caption{Conflict-free Success Rate and Allocation Time.}
\label{tab:success_rate}
\begin{tabular}{|c|c|c|c|c|c|}
\hline
\textbf{Metric} & \textbf{Method} & \textbf{4} & \textbf{8} & \textbf{12} & \textbf{20} \\
\hline
\multirow{3}{*}{\textbf{Success Rate (\%)}} & Hungarian & 100 & 100 & 100 & 100 \\
&MAGNNET & 100 & 100 & 90 & 80 \\
&Greedy & 90 & 80 & 80 & 60 \\
\hline
\multirow{3}{*}{\textbf{Allocation Time (s)}}&Hungarian & 0.8 & 1.5 & 2.8 & 5.6 \\
&MAGNNET & 0.4 & 0.6 & 1.2 & 2.8 \\
&Greedy & 0.3 & 0.3 & 0.5 & 1.2 \\
\hline
\end{tabular}
\end{table}

Table~\ref{tab:success_rate} shows the conflict-free success rates against baseline Hungarian and greedy approaches.
While the Hungarian method requires increased computation time with higher agent counts, our MAGNNET remains more scalable and maintains quick decision times. The Greedy method is faster but exhibits lower success rates.

\subsubsection{Continuous task generation}
In a more realistic scenario, tasks arrive dynamically over time. Our approach adapts to ongoing task announcements by letting each agent independently decide acceptance or rejection. Results indicate that, with additional training steps and an appropriately tuned reward structure, the MAGNNET approach mitigates conflicts while retaining near-optimal allocations and can handle continuous task generations. %Occasional conflicts still arise in rapidly changing environments, suggesting the need for further reward shaping or improved communication strategies.

\section{Conclusion and Future Work}
\label{sec:conclusion}
In this paper, we introduce a decentralized task allocation framework for multi-agent autonomous vehicle systems MAGNNET, that integrates GNN and DRL in a CTDE paradigm. Simulation results demonstrate that our approach achieves near-optimal performance compared to a centralized Hungarian baseline and surpasses the baseline greedy method in conflict avoidance and total travel cost, especially under growing agent populations. Our method achieved 92.5\% success rate in assignment of conflict-free allocation, outperforming the baseline optimal and greedy approaches. Moreover, MAGNNET assigns tasks faster than the Hungarian algorithm. 

Future directions include transferring the trained policy to physical fleets of drones and ground robots, accounting for sensor noise, real-time communication delays, and unpredictable obstacles. In real world, this can be utilized using peer-to-peer ESP-NOW communication protocol. Further refining the reward structure and introducing an attention mechanism to improve conflict resolution in highly dynamic environments will be additionally applied. 

% Extending the framework to hierarchical task allocation in which tasks are composed of subtasks, potentially leveraging more advanced GNN architectures. Investigating scalable attention mechanisms or bandwidth-aware message scheduling that reduce overhead in larger networks and also incorporating the MAGNNET to Taxi matching problem that solves for autonomous cars.
% Overall, our approach serves as a promising step toward robust, conflict-free task allocation in multi-robot systems where scalability and real-time autonomy are paramount.

% \bibliographystyle{IEEEtran}
\bibliography{references}

\end{document}